\documentclass[journal]{IEEEtran}
\IEEEoverridecommandlockouts                              
\usepackage{lipsum}
\usepackage{colortbl}
\usepackage{url}
\usepackage{graphicx}
\usepackage{amsmath}
\usepackage{amssymb}
\usepackage{multirow}
\usepackage{array}
\usepackage[linesnumbered,ruled,vlined]{algorithm2e}
\usepackage{booktabs}
\usepackage[noadjust]{cite}
\usepackage{soul}
\usepackage[colorlinks,allcolors=black]{hyperref}

\usepackage{multirow}
\usepackage[table,xcdraw]{xcolor}

\ifCLASSINFOpdf
\else
\fi
\ifCLASSOPTIONcompsoc
\else
  \usepackage[caption=false,font=footnotesize]{subfig}
\fi
\begin{document}
\newtheorem{assumption}{Assumption}

\title{Simulation-to-reality UAV Fault Diagnosis with Deep Learning}
%
%
%

\author{Wei Zhang, Junjie Tong, Fang Liao and Yunfeng Zhang
\thanks{ \textit{(Corresponding author: Junjie Tong.)}}
\thanks{Wei Zhang, Junjie Tong and Yunfeng Zhang are with the Department of Mechanical Engineering, National University of Singapore.
	{\tt\small e-mail: weizhang@u.nus.edu,tongjj@nus.edu.sg, mpezyf@nus.edu.sg}
	}%
\thanks{Fang Liao are with Temasek Laboratories, National University of Singapore, Singapore.
	{\tt\small e-mail:tsllf@nus.edu.sg}
	}%
}
\markboth{Preprint.}
{Wei Zhang \MakeLowercase{\textit{et al.}}: IPAPRec: A promising tool for learning high-performance mapless navigation skills with deep reinforcement learning} 
\maketitle
\begin{abstract}
Accurate diagnosis of propeller faults is crucial  for ensuring the safe and efficient operation of quadrotors. Training a fault classifier using simulated data and deploying it on a real quadrotor is a cost-effective and safe approach. However, the simulation-to-reality gap often leads to poor performance of the classifier when applied in real flight. In this work, we propose a deep learning model that addresses this issue by utilizing newly identified features (NIF) as input and utilizing domain adaptation techniques to reduce the simulation-to-reality gap. In addition, we introduce an adjusted simulation model that generates training data that more accurately reflects the behavior of real quadrotors. The experimental results demonstrate that our proposed approach achieves an accuracy of 96\% in detecting propeller faults. To the best of our knowledge, this is the first reliable and efficient method for simulation-to-reality fault diagnosis of quadrotor propellers.
\end{abstract}


%
\IEEEpeerreviewmaketitle

\section{Introduction}
\begin{figure}[htp]
	\centering
	\includegraphics[width=0.8\linewidth]{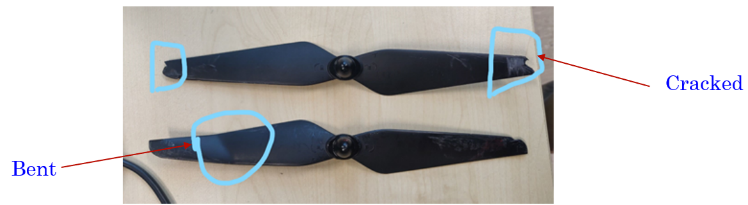}
	\caption{Examples of broken propellers}
	\label{propeller}
\end{figure}

%
%
%
%
\IEEEPARstart{Q}{uadrotors} , also known as quadrotor unmanned aerial vehicles (UAVs), are increasingly being utilized for a variety of applications, including search-and-rescue, homeland security, package delivery, and military surveillance \cite{shraim2018survey}. However, the propellers of these UAVs may be subject to damage during task execution (see Fig. \ref{propeller}) due to unexpected circumstances, such as collision with obstacles. Such damage can compromise the ability of the UAV to complete its assigned tasks, and may even result in catastrophic damage to the UAV itself and potentially those in the immediate vicinity. To mitigate these risks, real-time monitoring of the condition of quadrotor propellers is crucial. By detecting and addressing issues with the propellers in a timely manner, we can ensure the safe and effective operation of these UAVs, and ultimately prevent potential damages and injuries.

Model-based methods are a common approach for detecting faults in quadrotors \cite{guzman2019actuator}, \cite{amoozgar2013experimental}. These methods involve first developing system models and then identifying faults based on the difference between the model output and the actual output. However, accurately modeling quadrotor systems can be challenging, as it requires extensive expert knowledge and may require multiple experiments to identify structural parameters.

In contrast, data-driven approaches, which operate in a model-free manner, have received increasing attention in recent years \cite{iannace2019fault}, \cite{park2022multiclass}. These approaches utilize deep neural networks (DNNs) \cite{hinton2006reducing} to directly map sensor input to the results of the diagnosis. The parameters of the DNNs are learned from real-world flying data. As they are data-driven and work in an end-to-end fashion, such approaches do not require the development of system models or the design of fault classifiers based on these models. Much of the previous work in this area \cite{yang2021intelligent}, \cite{yang2021intelligentb} has focused on designing powerful DNN models that can learn the complex nonlinear relationship between the input features and use this knowledge to identify propeller faults.

While data-driven approaches have demonstrated promising results for fault diagnosis in quadrotors, collecting flying data with broken propellers can be dangerous and costly. In contrast, generating fault data in simulation is a safer and more cost-effective option. However, there is often a discrepancy between simulation data and real flight data, known as the simulation-to-reality gap, which can result in poor performance of classifiers trained on simulation data when applied to real-flight data.

In this research, we focus on improving the performance of the classifier in the real-world by reducing the simulation-to-reality gap. To achieve this goal, we propose newly identified features (NIF) that exhibit a linear relationship between input features, allowing for efficient learning by the DNN model. These features are also capable of generalizing to data from different domains. In addition, we utilize a domain adaptation (DA) method to further enhance the cross-domain performance of the classifier. Finally, we propose an adjusted simulation model to generate training data that more accurately reflects the unbalanced behavior of real quadrotors. Overall, the contributions of our work include:

\begin{itemize}
	\item Newly identified features are proposed to enhance the robustness of the classifier to domain changes.
    \item Domain adaptation is utilized to reduce the domain gap and improve the performance of the classifier in the target domain.
    \item With the help of NIF, DA and the adjusted simulation model, the fault classifier trained with simulation data can achieve high accuracy when applied to real flight.
\end{itemize}

The remainder of this paper is organized as follows. A brief overview of related works is provided in Section \ref{RW}, followed by the problem description in Section \ref{Preliminaries}. The intelligent diagnosis method is introduced in Section \ref{approach}, and extensive experiments are conducted to evaluate our method in Section \ref{Implementation}. Last, we draw the conclusions and the future works in Section \ref{Conclusion}.

\section{Related works}\label{RW}
To identify faults in propellers using deep learning, there are two main approaches based on the input source. The first involves using a specific sensor to monitor propeller conditions, with microphone arrays being a popular choice. For example, Katta et al. \cite{katta2022real} used microphone arrays to detect propeller faults and released a dataset for further research. They first converted audio data into Mel-frequency cepstral coefficient (MFCC) features and used deep neural networks such as convolutional neural networks (CNNs) \cite{lecun1995convolutional} 
and transformer \cite{vaswani2017attention} encoders as fault classifiers. Similarly, Liu et al. \cite{liu2020audio} used CNNs to diagnose faults from the time-frequency spectrum of audio signals, and also employed transfer learning to reduce the training time of fault classifiers for quadrotors with different dynamics. Instead of installing microphones on the quadrotor, Iannace et al. \cite{iannace2019fault} placed microphones around the quadrotor to detect unbalanced propellers. With this setup, a one-hidden-layer neural network achieved an accuracy of 97.63\%. While this approach has high accuracy in indoor scenarios, its performance in outdoor environments, particularly under windy conditions, may be impaired. Additionally, it requires installing a microphone array on top of the quadrotor, which increases cost and battery consumption.

The second approach involves using flying data to monitor propeller conditions. Yang et al. \cite{yang2021intelligent} proposed a deep residual shrinkage network as a classifier using quadrotor state information (roll value, pitch value, roll rate, pitch rate, and yaw rate) and the four angular velocities of the propeller as input. Additionally, Park et al. \cite{park2022multiclass} used a stacked pruning sparse denoising autoencoder to process image inputs consisting of 20 signals with a length of 20 containing drone attitude, position, velocity, and acceleration information, and employed a CNN as a classifier. This approach exhibits good performance in noisy environments due to the denoising operation. However, although the trained classifiers demonstrated high accuracy during testing, their performance in different scenarios is unknown.

\section{Preliminaries}\label{Preliminaries}
First of all, abbreviations which are frequently used in this work are summarized in Table \ref{Abbreviations}.

\begin{table}[htp]
\caption{\label{Abbreviations} Abbreviations and descriptions}
\centering
\begin{tabular}{ll}
\hline
\hline
Abbreviation & Description                                   \\ \hline
CF           & Conventional features                         \\
DA           & Domain adaptation                             \\
DCNN         & Deep convolutional   neural network           \\
MMD          & Maximum Mean   Discrepancy                    \\
NIF          & Newly identified   features                   \\
t-SNE        & t-distributed   stochastic neighbor embedding \\ \hline
\hline
\end{tabular}
\end{table}

\subsection{Problem description}
The UAV fault diagnosis problem can be approached as a classification task, with the goal of accurately identifying the fault type of a given UAV based on its flying data. As shown in Table \ref{labels}, there are five potential fault categories, including an all-healthy category and four faulty categories. To train a classifier, we collect and label flying data for each of these categories. Utilizing this labeled dataset, we can then train a deep neural network as a classifier, with the aim of accurately predicting the fault type of a given UAV based on its flying data. The classifier has five outputs, each corresponding to one of the five fault categories.

\begin{figure}[htp]
	\centering
	\includegraphics[width=1.0\linewidth]{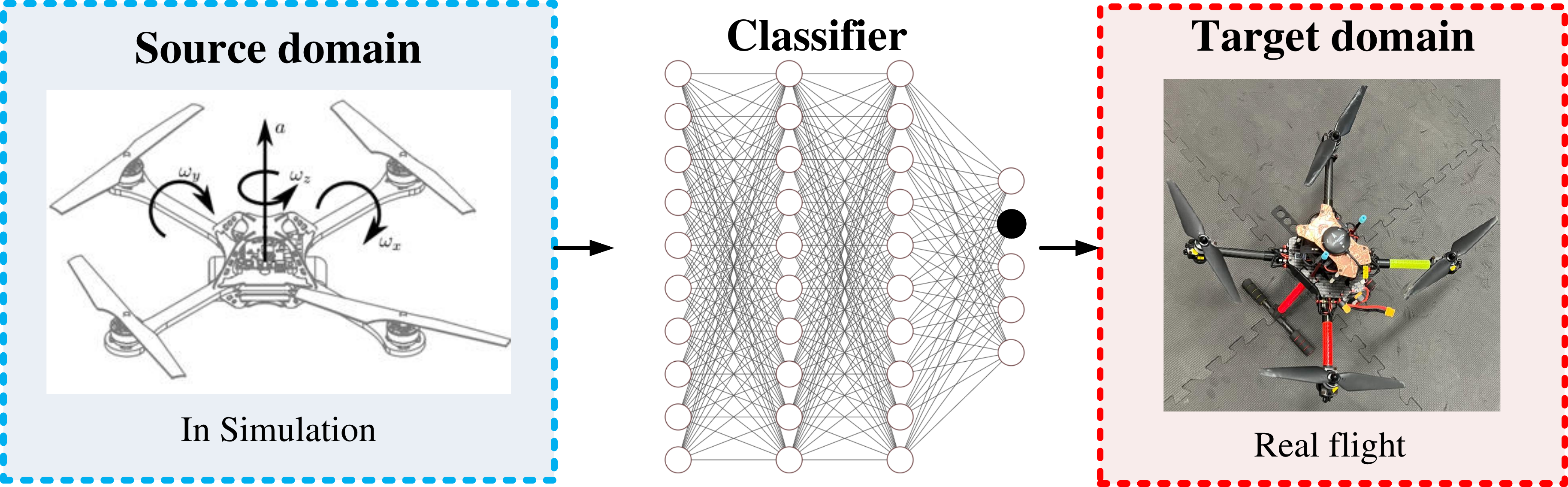}
	\caption{Illustration of the simulation-to-reality UAV fault diagnosis problem.}
	\label{prod}
\end{figure}

In this study, as shown in Fig. \ref{prod}, we aim to develop a fault classifier for UAV propeller using simulated data, and subsequently apply it to real-world fault detection. This simulation-to-reality UAV fault diagnosis problem can be modeled as a cross-domain classification problem, with the simulation data serving as the source domain and the real-world data serving as the target domain.

One challenge in this scenario is that the simulation model may not perfectly match the real-world counterpart, due to factors such as noise and installation error. As a result, a classifier trained solely on source domain data may perform poorly when applied to the target domain. To reduce the domain gap, during training, the classifier also has access to all-healthy samples from the target domain.  The objective of this paper is to learn a fault classifier with high fault-classification accuracy in target domain.

\begin{table}[htp]
\caption{\label{labels} Labels of five fault categories.}
\centering
\begin{tabular}{lllll}
\hline
\hline
label` & Propeller 1 & Propeller 2 & Propeller 3 & Propeller 4 \\ \hline
1 & healthy & healthy & healthy & healthy \\
2 & faulty & healthy & healthy & healthy \\
3 & healthy & faulty & healthy & healthy \\
4 & healthy & healthy & faulty & healthy \\
5 & healthy & healthy & healthy & faulty \\ 
\hline
\hline
\end{tabular}
\end{table}

\section{APPROACH}\label{approach}
\begin{figure}[htp]
	\centering
	\includegraphics[width=1.0\linewidth]{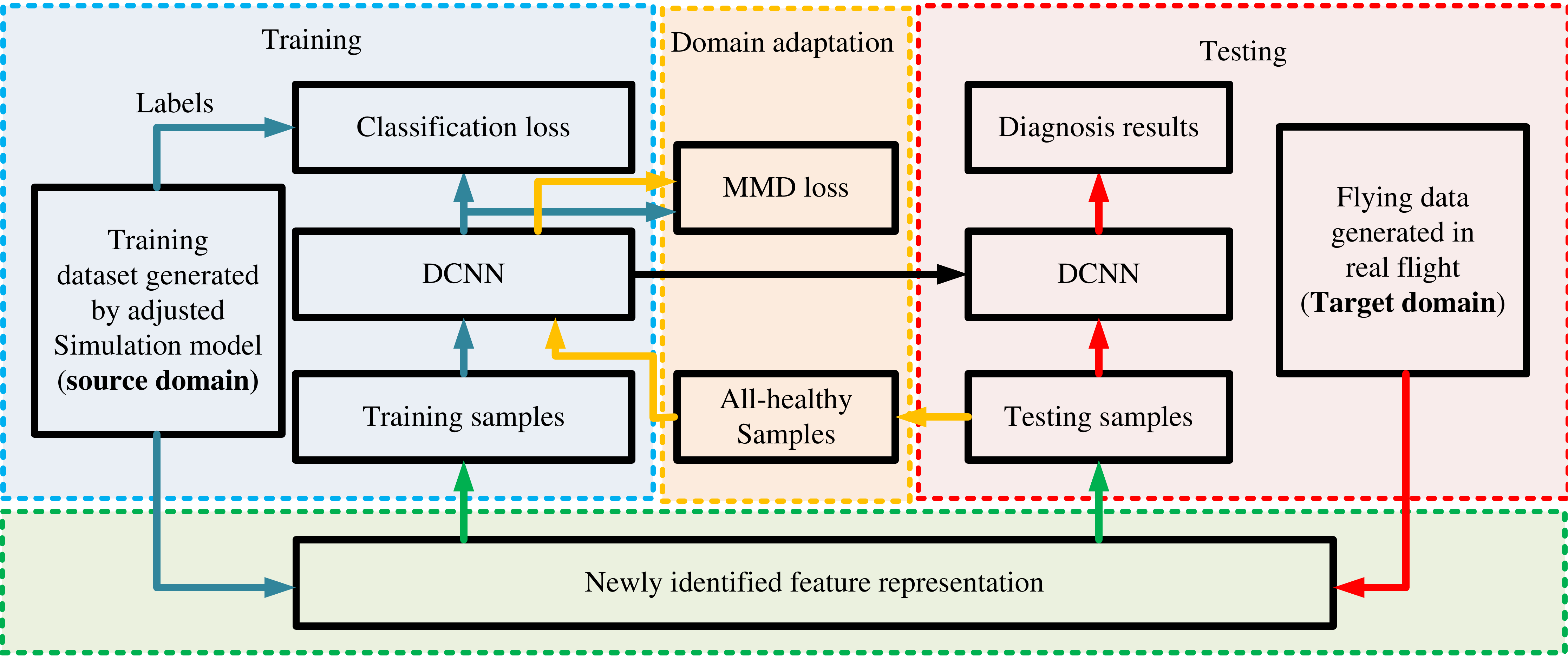}
	\caption{Framework of the proposed approach.}
	\label{Framework}
\end{figure}

In this research, we propose a novel approach for monitoring the health status of quadrotor propellers using flying data. The comprehensive framework of the proposed approach is depicted in Fig. \ref{Framework}. Both training and testing data use the proposed newly identified input representation. The end-to-end model employed in the training process is a DCNN (Deep Convolutional Neural Network) \cite{krizhevsky2017imagenet}, which utilizes multi-source features as input and contains multiple convolutional layers.  The data used for training is generated by the adjusted simulation model. During the testing phase, the trained DCNN model can be utilized for the diagnosis of propeller faults in real-world scenarios. 
\subsection{Newly identified input representation}\label{A1}
In this paper, onboard flying data are utilized to monitor the health status of the quadrotor propeller. In selecting the input for the DCNN, we consider two main criteria. First, the input should contain both the input and output information for the propeller. This is because, when a propeller is damaged, the output of the propeller will change in response to the same input information. By analyzing these changes, the DCNN can effectively detect faults or failures. Second, we aim to minimize the complexity of the relationship between the input and output information. A linear relationship is generally the most straightforward and easiest for the DCNN to process, as nonlinear relationships may be more difficult for the network to effectively extract useful features from the input representation.

\begin{figure}[htp]
	\centering
	\includegraphics[width=0.8\linewidth]{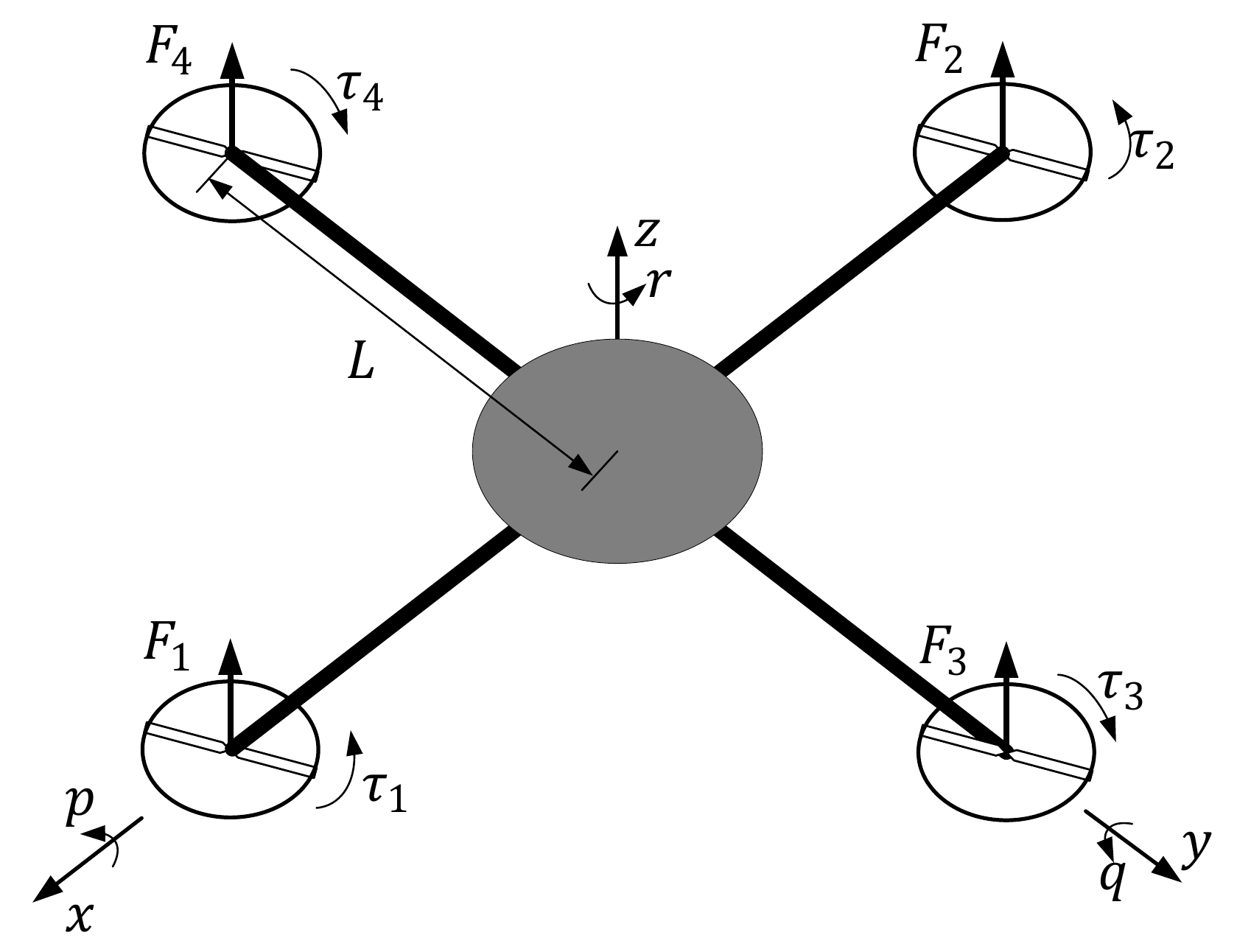}
	\caption{The configuration of the quadrotor in the body frame.}
	\label{UAV}
\end{figure}

Based on these considerations, we select DCNN input from the quadrotor dynamic equation. By doing so, we are able to effectively capture both the input and output information for the propeller, as well as maintain a relatively simple relationship between these two variables. This allows for accurate and efficient fault detection, ultimately contributing to the overall health and safety of the quadrotor system. As shown in Fig. \ref{UAV}, based on Euler’s Equations of Motion, the relationship between the quadrotor angular accelerations $\left(\dot{p},\dot{q},\dot{r}\right)$, the forces $F_{i=1,2,3,4}$ and the torques $\tau_{i=1,2,3,4}$ of the four propellers is given in Equation \ref{dy_eq}. 

\begin{equation}\label{dy_eq}
\begin{aligned}
I\left[\begin{matrix}\dot{p}\\\dot{q}\\\dot{r}\\\end{matrix}\right]=\left[\begin{matrix}L\left(F_3-F_4\right)\\L\left(F_2-F_1\right)\\\tau_1-\tau_3+\tau_2-\tau_4\\\end{matrix}\right]-\left[\begin{matrix}p\\q\\r\\\end{matrix}\right]\times I\left[\begin{matrix}p\\q\\r\\\end{matrix}\right].
\end{aligned}
\end{equation}
where $I$ denotes the inertia matrix as follows,
\begin{equation}\label{inertia}
\begin{aligned}
I=\left[\begin{matrix}I_{xx}&0&0\\0&I_{yy}&0\\0&0&I_{zz}\\\end{matrix}\right]
\end{aligned}
\end{equation}
where $I_{xx}$, $I_{yy}$ and $I_{zz}$ are the moments of inertia along the $x$, $y$, and $z$ axes of the quadrotor. Since $F_i=k_F\omega_i^2$ and $\tau_i=k_\tau\omega_i^2$, where $k_F$ and $k_\tau$ are two constants, Equation \ref{dy_eq} becomes
\begin{equation}\label{dy_eq2}
\begin{aligned}
I\left[\begin{matrix}\dot{p}\\\dot{q}\\\dot{r}\\\end{matrix}\right]=\left[\begin{matrix}Lk_F\left(\omega_3^2-\omega_4^2\right)\\Lk_F\left(\omega_2^2-\omega_1^2\right)\\k_\tau\left(\omega_1^2-\omega_3^2+\omega_2^2-\omega_4^2\right)\\\end{matrix}\right]-\left[\begin{matrix}p\\q\\r\\\end{matrix}\right]\times I\left[\begin{matrix}p\\q\\r\\\end{matrix}\right]
\end{aligned}
\end{equation}
When one of the propellers breaks, under the same rotational speed of the motor, the generated force and torque are different from the healthy counterpart. Once the force and torque of the propeller change, the quadrotor’s angular accelerations will change accordingly. Based on this observation and our two criteria, at time step $t$, the input $x_t$ to the DCNN is designed as as follows,
\begin{equation}\label{input_eq}
\begin{aligned}
x_t=\left[\begin{matrix}{\dot{p}}_{t-T}&{\dot{p}}_{t-T+1}&\cdots&{\dot{p}}_t\\{\dot{q}}_{t-T}&{\dot{q}}_{t-T+1}&\cdots&{\dot{q}}_t\\{\dot{r}}_{t-T}&{\dot{r}}_{t-T+1}&\cdots&{\dot{r}}_t\\\omega_{1, t-T}^2&\omega_{1, t-T+1}^2&\cdots&\omega_{1, t}^2\\\omega_{2, t-T}^2&\omega_{2, t-T+1}^2&\cdots&\omega_{2, t}^2\\\omega_{3, t-T}^2&\omega_{3, t-T+1}^2&\cdots&\omega_{3, t}^2\\\omega_{4, t-T}^2&\omega_{4, t-T+1}^2&\cdots&\omega_{4, t}^2\\\end{matrix}\right].
\end{aligned}
\end{equation}
where $T$ ($T=80$ in this paper) is the length of the time window, which controls the range of historical information of the input. As shown, $x_t$ can be treated as a temporal signal which contains seven feature channels. As these input features are newly identified by this work, we refer to $x_t$ as newly identified features (NIF).
\subsection{DCNN model with domain adaptation}\label{AdaIP}
Recently, deep convolutional neural networks (DCNNs) have gained significant traction as a powerful end-to-end classification tool for fault diagnosis tasks \cite{Zhang2017}, \cite{Zhang2018}. These networks are able to learn to extract useful features from multi-channel time-sequence input data through the use of convolutional layers, and subsequently classify these features using dense layers.

\begin{figure}[htp]
	\centering
	\includegraphics[width=0.9\linewidth]{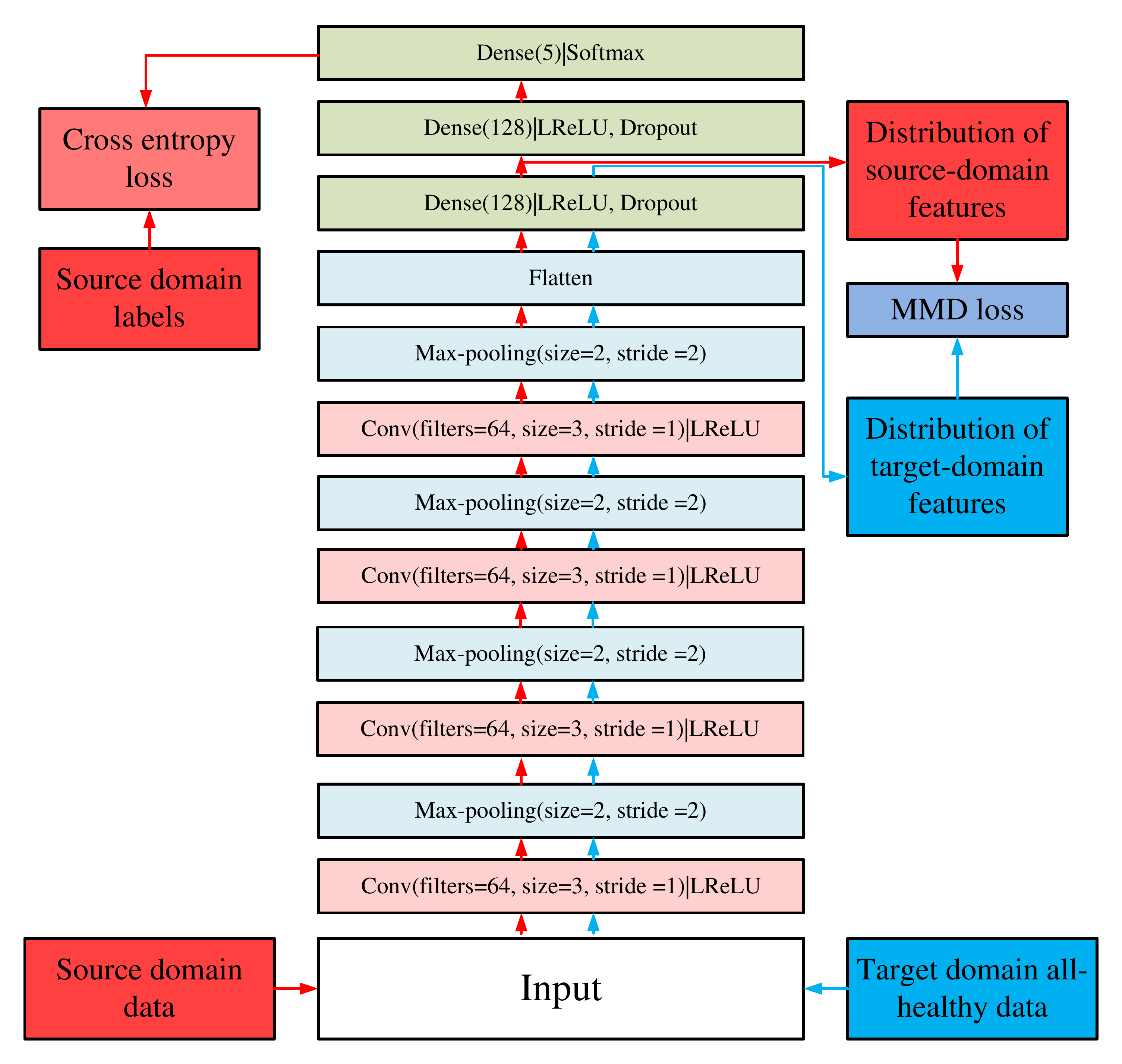}
	\caption{The structure of DCNN.}
	\label{DCNN}
\end{figure}

The structure of the DCNN model used in this study is depicted in Fig. \ref{DCNN}. The input of DCNN (see Eq. \ref{input_eq}) is treated as a one-dimensional signal with seven input channels. As shown, the first layer is a 1D convolutional layer with a kernel size of three and 64 filters. Following the convolutional layer, a max-pooling operation is performed to reduce the dimensions of the extracted features. This convolutional and max-pooling module is repeated three times in subsequent layers, with the goal of effectively extracting useful feature representations from the input data.

After the four convolutional and max-pooling modules, two dense layers with 128 neurons and a softmax layer are employed to classify the extracted features into five distinct categories. In order to enhance the generalization performance of the model, layer dropout \cite{Srivastava2014} with a dropout rate of 0.1 is applied to the two dense layers. The output layer utilizes the Softmax function to transform the logits of the five neurons into discrete probability distributions representing the five possible quadrotor health conditions. The formula for the Softmax function is as follows:
\begin{equation}\label{dynamic}
\begin{aligned}
q\left(z_j\right)=\text{softmax}\left(z_j\right)=\frac{e^{z_j}}{\sum_{i=1}^{5}e^{z_i}}.
\end{aligned}
\end{equation}
where $z_j$ denotes the logits of the $j$-th output neuron.

The classification loss of our DCNN model is the cross-entropy loss between the estimated softmax output probability distribution and the target class probability distribution. Let $p(x)$ denote the target distribution (the one-hot format labels) and $q(x)$ denote the estimated distribution outputted by DCNN, the cross-entropy between $p(x)$ and $q(x)$ is:
\begin{equation}\label{dynamic}
\begin{aligned}
\mathcal{L}_{cross_entropy}= H(p(x),q(x))=-\Sigma_{x}p(x)\log q(x).
\end{aligned}
\end{equation}

When a fault classifier is utilized for fault diagnosis, its performance may be compromised due to the different distributions of the target and source domain data. To address this issue, domain adaptation (DA) methods are frequently employed to align the distribution of the target domain data with that of the source domain data. However, typical DA approaches require access to the full range of unlabeled data from the target domain, which is not feasible in our case due to the risks associated with flying a quadrotor with a broken propeller. Therefore, our DA method aims to utilize only the all-healthy samples from the target domain to extract transferable features, which allows for a safer approach. Similar to \cite{Wen2018}, only the all-healthy samples from the target domain are utilized for DA. Our DA method is under the following assumption,

\begin{assumption}\label{assumption1}
The cross-domain variation way for samples of different faults is similar
\end{assumption}

Under the assumption that the distributions of healthy source and target domain data are brought closer together, it is possible to decrease the distance between the source and target domain distributions of other categories. In this paper, we use Maximum Mean Discrepancy (MMD) to measure the distance between the distributions of all-healthy data in the source and target domains in the feature space. As shown Fig. \ref{DCNN}, the features used for calculating the MMD loss are extracted from the first dense layer. With the features, the MMD loss is defined as follows,
\begin{equation}\label{MMD loss}
\begin{aligned}
\mathcal{L}_{MMD}=\lVert\frac{1}{n_{c}^s}\sum_{p=1}^{n_{c}^s}\phi\left(x_{c,p}^s\right)-\frac{1}{n_{\omega_c}^t}\sum_{q=1}^{n_{c}^t}\phi\left(x_{c,q}^t\right)\rVert_\mathcal{H}^2.
\end{aligned}
\end{equation}
where $\phi(x)$, representing the features outputted by the first fully-connected layer, and $x_{c}^s$ and $x_{c}^t$ are samples of the $c$ ($c=1$ in this report) category from $\mathcal{D}^s$ and $\mathcal{D}^t$, respectively. The minimization of the MMD loss serves to align the distribution of real flight all-healthy samples with that of simulated all-healthy samples.

By integrating the classification loss and MMD loss together, the model loss function is as follows,
\begin{equation}\label{dynamic}
\begin{aligned}
\mathcal{L} = \mathcal{L}_{cross\_entropy}+\lambda\mathcal{L}_{MMD}
\end{aligned}
\end{equation}
where $\lambda$ is a constant which weighs the contribution of the MMD loss.

\subsection{Adjusted simulation model}
In our previous simulation model \cite{tong2023machine}, we follow an ideal dynamic model of the quadrotor, i.e., if given four propellers with the same rotation speed, it should be in an upright hovering position with angular velocity and acceleration equal to zero. However, in the real world, this behavior is often not realistic because of the following  factors: 1) the center of gravity that is off-centered; 2) the imperfect condition of the mounted propeller or motor; 3) the inaccuracy of flight sensors as well as the imperfect build of the quadrotor’s structure. Therefore, to make our simulation more accurate in fault detection on real flight data, we need to adjust our simulation model based on the real-flight data to account for this phenomenon.

To achieve this, the rotational speeds of the motors are adjusted by multiplying the unbalanced ratio $\rho$. The unbalanced ratio is designed by the following observation: if the average speed of motor \textit{i} is higher than motor 1 at steady state, which means that, on average, motor \textit{i} must rotate faster to keep the quadrotor in a stable position. Hence, in simulation, motor \textit{i} is assumed to be weaker than motor 1 such that it is required to rotate at a faster rate to produce similar torque and thrust. The unbalanced ratio can be computed using the average RPM ($\bar\omega_i$) of each respective motor over the baseline motor 1 ($\bar\omega_1$) during steady state, i.e.,
\begin{equation}
    \rho_i=\frac{\bar\omega_i}{\bar\omega_1}
\end{equation}
where \textit{i} is index of the \textit{i-}th motor. As $\rho_i$ varies with $\omega_i$, a linear interpolation approach is used to compute the specific ratio at each respective $\omega_i$ value of the motor. The computation of $\rho_i(\omega_i)$, is as follows,
\begin{equation}
    \rho_i(\omega_i)=\frac{\omega_i}{\max(\omega_i)}\rho_i
\end{equation}
with the specific unbalanced ratio $\rho_i(\omega_i)$, the adjusted rotational speed of motor \textit{i} is as follows
\begin{equation}
    \omega_{i, a}=\rho_i(\omega_i)\omega_i
\end{equation}

\section{Implementation and test}\label{Implementation}
In this section, we evaluate the simulation-to-reality performance of the proposed method. The training data for the DCNN model is generated using the adjusted simulation model. After training, the DCNN model is applied to classify faults on real flight data."

\subsection{Dataset Description}
The quadrotor angular accelerations $\left(\dot{p_t},\dot{q_t},{\dot{r}}_t\right)$ can be calculated using the angular velocities of the quadrotor. However, the motor velocities cannot be directly measured or calculated using the available flying data. According to \cite{Powers2015}, the relationship between the real motor velocity $\omega_i$ and the commanded motor velocity $\omega_i^{com}$ is as follows,
\begin{equation}\label{dynamic}
\begin{aligned}
{\dot{\omega}}_i=k_m\left(\omega_i^{com}-\omega_i\right).
\end{aligned}
\end{equation}
where the time constant $k_m$ is the motor gain. As a first approximation, it can be assumed that the motor controller is perfect and that the time constant related to the motor response is arbitrarily small \cite{Powers2015}. In other words, it can be assumed that the real motor velocity and the commanded motor velocity are equal. Hence, in this paper, we use $\omega_i^{com}$ as the $\omega_i$ for fault diagnosis. 

\begin{figure}[htpb]
    \centering
	  \subfloat[Data collection in simulation]{
       \includegraphics[width=0.45\linewidth]{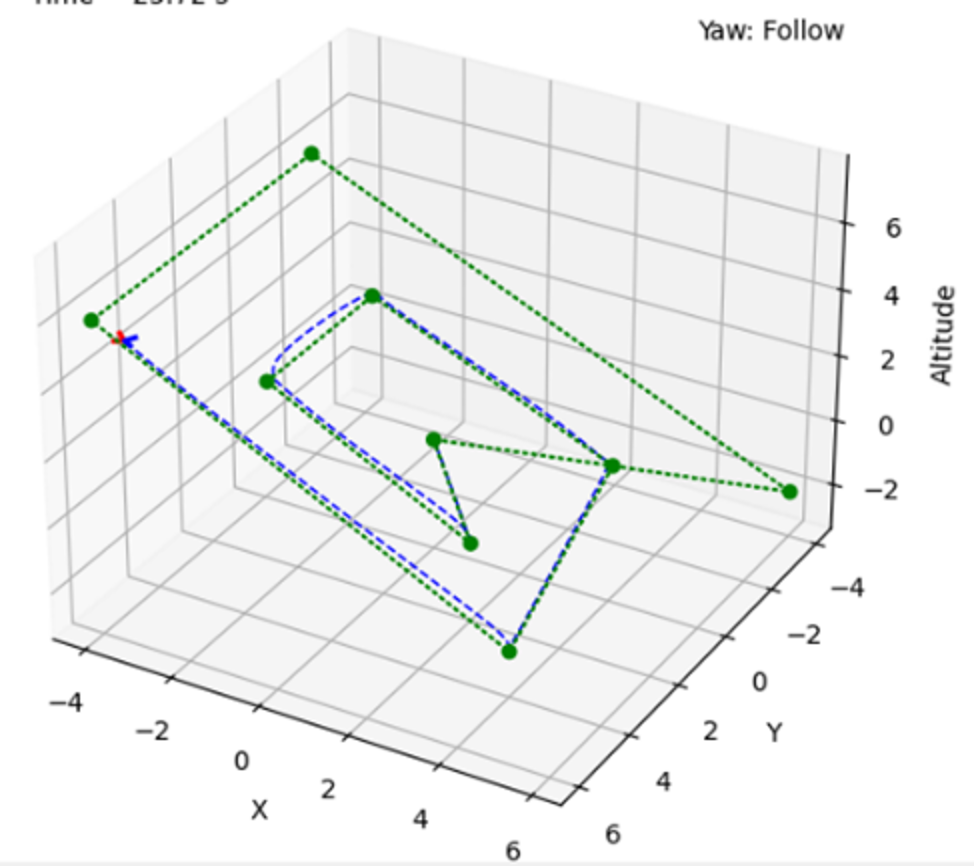}\label{sim_exp}}
	  \subfloat[Data collection in real flight]{
        \includegraphics[width=0.45\linewidth]{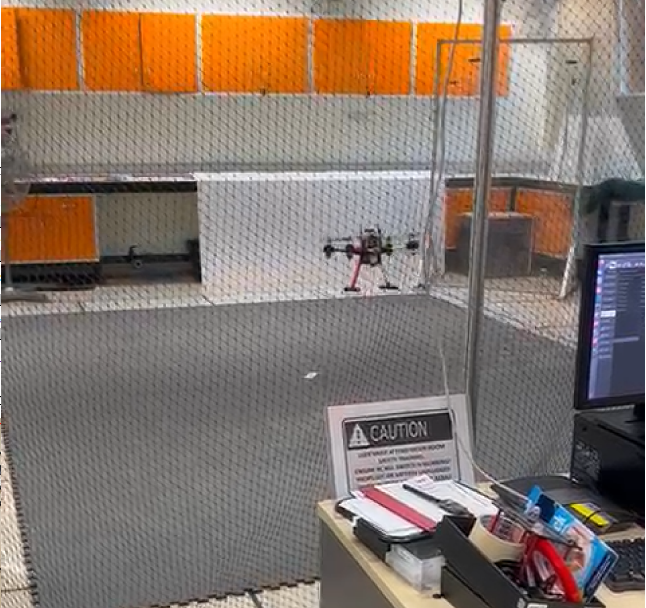}\label{real_exp}}
	\caption{Data collection in simulation and real flight.}
	\label{data_collection}
\end{figure}

The adjusted simulation model was utilized to generate dataset S, which will be employed for the training of the DNN model. As illustrated in Fig. \ref{sim_exp}, for each fault category, the UAV autonomously flew to the specified waypoints during data collection in the simulation process. As depicted in Table \ref{real_dataset}, a total of 800 samples were generated for each fault category using the adjusted simulation model, resulting in a comprehensive dataset for each fault category.

\begin{table*}[htp]
\centering
\caption{\label{real_dataset} Description of quadrotor datasets generated from simulation (for training) and real flight (for testing).}
\begin{tabular}{|c|c|c|ccccc|}
\hline
\multirow{3}{*}{\begin{tabular}[c]{@{}c@{}}Dataset\\ name\end{tabular}} & \multirow{3}{*}{\begin{tabular}[c]{@{}c@{}}Dataset type\end{tabular}} & \multirow{3}{*}{\begin{tabular}[c]{@{}c@{}}Dataset domain\end{tabular}} & \multicolumn{5}{c|}{Category label (fault location)} \\ \cline{4-8} 
 &  &  & \multicolumn{1}{c|}{1 (None)} & \multicolumn{1}{c|}{2 (Propeller 1)} & \multicolumn{1}{c|}{3 (Propeller 2)} & \multicolumn{1}{c|}{4 (Propeller 3)} & 5 (Propeller 4) \\ \cline{4-8} 
 &  &  & \multicolumn{5}{c|}{Number of samples} \\ \hline
S & Training & \begin{tabular}[c]{@{}c@{}}Source domain (Simulation)\end{tabular} & \multicolumn{1}{c|}{800} & \multicolumn{1}{c|}{800} & \multicolumn{1}{c|}{800} & \multicolumn{1}{c|}{800} & 800 \\ \hline
T & Testing & \begin{tabular}[c]{@{}c@{}}Target domain  (Real flight)\end{tabular} & \multicolumn{1}{c|}{800} & \multicolumn{1}{c|}{800} & \multicolumn{1}{c|}{800} & \multicolumn{1}{c|}{800} & 800 \\ \hline
\end{tabular}
\end{table*}

\begin{figure}[htpb]
    \centering
	  \subfloat[Broken propellers]{
       \includegraphics[width=0.7\linewidth]{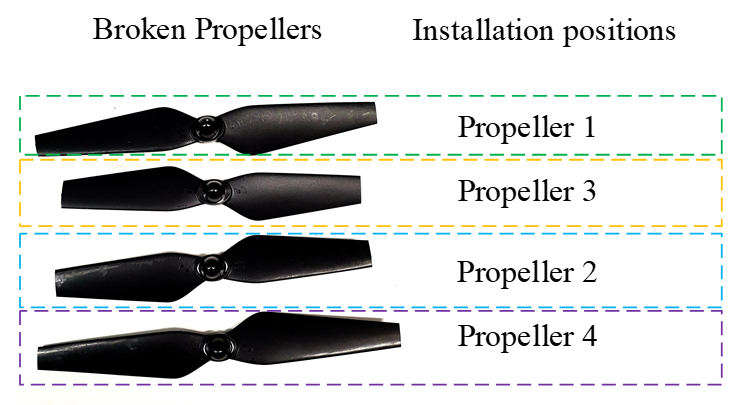}\label{bp}}
        \\
	    \subfloat[Real UAV model]{
        \includegraphics[width=0.7\linewidth]{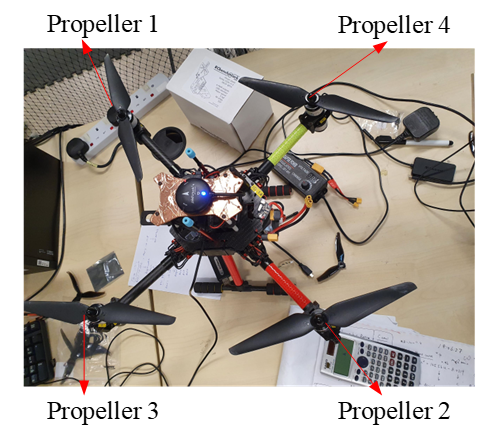}\label{real_UAV}}
	\caption{The real UAV model and broken propellers used in real-world data collection.}
	\label{real_propeller}
\end{figure}

In order to gather real flight data for testing purposes, indoor experiments were conducted as depicted in Fig. \ref{real_exp}. Prior to the real flight, the real UAV was equipped with a broken propeller (see Fig \ref{bp}) in the designated position (see Fig. \ref{real_UAV}). It is important to note that only one broken propeller was installed for each fault category, while the remaining propellers were healthy. During data collection, the UAV was manually controlled by a pilot for approximately two minutes, and this procedure was repeated five times to encompass the five fault categories. The resulting real flight data was used to generate Dataset T for testing purposes.

\subsection{Training details}
To assess the performance enhancement of the proposed NIF on the DCNN model in the target domain, a comparison was conducted with conventional features (CF, as used in \cite{yang2021intelligent}). Similar to NIF, CF also comprises a multi-channel temporal signal with nine feature channels: roll value, pitch value, roll rate, pitch rate, yaw rate, and the four rotational speeds of the propellers. As CF has nine feature channels, the channel number of its DCNN model is nine. Apart from that, the network structures of the DCNN models utilized for NIF and CF are equivalent. These models are referred to as DCNN+NIF and DCNN+CF, respectively. Furthermore, domain adaptation was applied to the DCNN+NIF model, resulting in the model named DCNN+NIF+DA. 

The training hyperparameters are presented in Table \ref{training}. The training process was terminated when the maximum number of training epochs was reached or when the training loss did not decrease after ten consecutive epochs. The utilization of a GPU (RTX 2080Ti) facilitated the acceleration of each training process, which took approximately ten minutes. The network saved for testing was the one with the smallest training loss. To evaluate the stability and repeatability of each model, the training process was repeated ten times.
\begin{table}[htp]
\caption{\label{training} Hyperparameter settings.}
\centering
{\begin{tabular}{ll}
\hline
\hline
Hyperparameters           & Value  \\ \hline
Mini-batch   size         & 128    \\
Learning   rate           & $5\times10^{-4}$ \\
Maximum   training epochs & 50     \\
Optimizer                 & Adam   \\
Dropout   rate            & 10\%  \\
Factor of MMD loss: $\lambda$            & $10^4$  \\
\hline
\hline
\end{tabular}}
\end{table}

\subsection{Results and comparison study}
\begin{figure}[htp]
	\centering
	\includegraphics[width=0.9\linewidth]{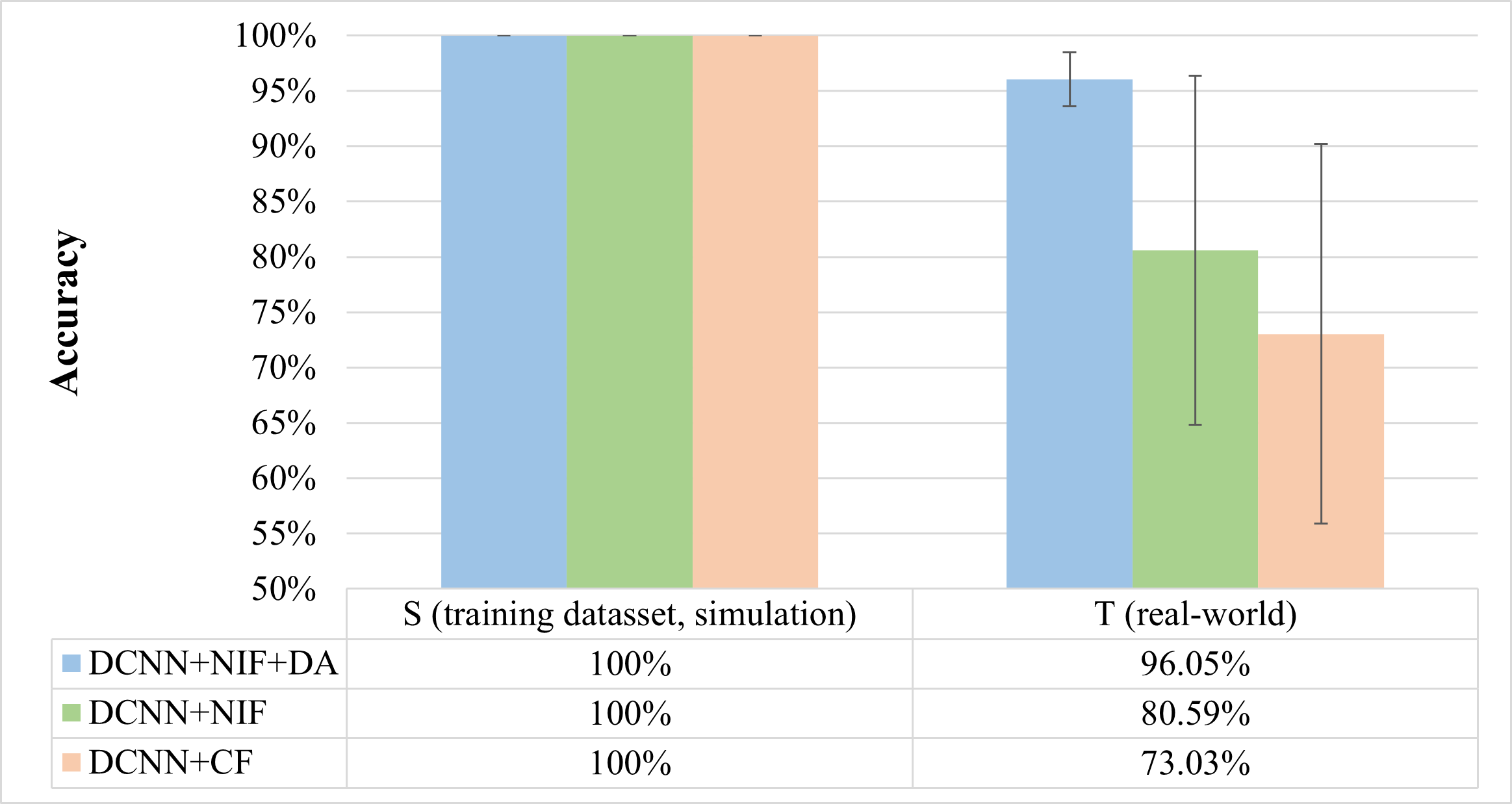}
	\caption{Accuracy of the three DCNN models on the Dataset S (simulation) and Dataset T (real flight).}
	\label{real_acc}
\end{figure}

After training, the DCNN model can be effectively utilized for fault classification tasks. The classification accuracy of the DCNN+CF, DCNN+NIF, and DCNN+NIF+DA models on Datasets S and T are presented in Fig. \ref{real_acc}. As depicted, the value of each bar represents the average accuracy over ten runs, with the standard deviation of the ten-time accuracy also displayed above each bar. It can be seen that the DCNN+CF model only attains an average accuracy of 73\%, while the proposed DCNN+NIF model achieves an accuracy of approximately 80\%, indicating NIF can enhance the generalization performance of the DCNN model. Moreover, the use of DA resulted in a significant enhancement in the accuracy of the DCNN model, with the accuracy increasing to 96\%. This demonstrates the reliability of the proposed DCNN+NIF+DA model as a fault classifier in real flight scenarios.

\begin{figure}[htp]
	\centering
	\includegraphics[width=0.9\linewidth]{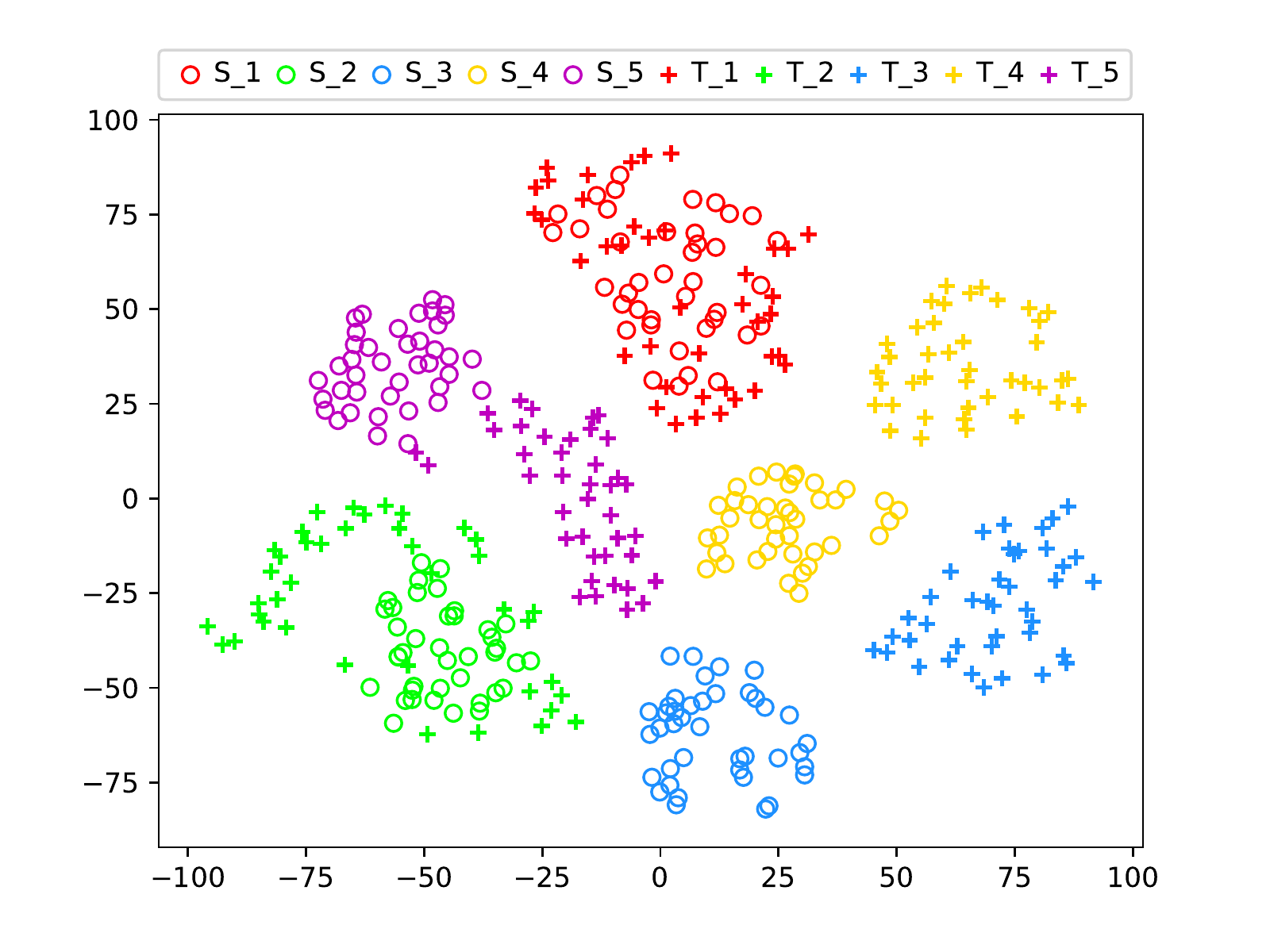}
	\caption{t-SNE visualization of features learned by DCNN+NIF+DA on Dataset S and Dataset T.}
	\label{real_tsne}
\end{figure}

Additionally, the features learned by the DCNN+NIF+DA model on Datasets S and T are compressed into two dimensions using t-SNE \cite{van2008visualizing}. The features are extracted by the first dense layer, which is also utilized in the calculation of the MMD loss. As shown in Fig. \ref{real_tsne}The distribution of target domain data is observed to be closely aligned with that of the source domain data, which may explain why the classifier trained on source domain data is able to make accurate predictions on the target domain data.


\section{Conclusion}\label{Conclusion}
In this paper, the proposed UAV fault diagnosis method effectively addresses the issue of poor performance of fault classifiers trained with simulated data due to the simulation-to-reality gap. By utilizing newly identified features and the MMD-based domain adaptation technique, as well as introducing the adjusted simulation model, our approach achieved an accuracy of 96\% in detecting propeller faults in real flight. This makes it the first reliable and efficient method for simulation-to-reality fault diagnosis of quadrotor propellers, ensuring the safe and efficient operation of these devices. In the future, we plan to extend the evaluation of our method to more challenging scenarios, such as windy outdoor environments. This will provide a better understanding of the robustness and limitations of our approach, and may inform further improvements.


%


\ifCLASSOPTIONcaptionsoff
  \newpage
\fi



%

\bibliographystyle{IEEEtran}
\bibliography{references}

\begin{thebibliography}{10}
\providecommand{\url}[1]{#1}
\csname url@samestyle\endcsname
\providecommand{\newblock}{\relax}
\providecommand{\bibinfo}[2]{#2}
\providecommand{\BIBentrySTDinterwordspacing}{\spaceskip=0pt\relax}
\providecommand{\BIBentryALTinterwordstretchfactor}{4}
\providecommand{\BIBentryALTinterwordspacing}{\spaceskip=\fontdimen2\font plus
\BIBentryALTinterwordstretchfactor\fontdimen3\font minus
  \fontdimen4\font\relax}
\providecommand{\BIBforeignlanguage}[2]{{%
\expandafter\ifx\csname l@#1\endcsname\relax
\typeout{** WARNING: IEEEtran.bst: No hyphenation pattern has been}%
\typeout{** loaded for the language `#1'. Using the pattern for}%
\typeout{** the default language instead.}%
\else
\language=\csname l@#1\endcsname
\fi
#2}}
\providecommand{\BIBdecl}{\relax}
\BIBdecl

\bibitem{shraim2018survey}
H.~Shraim, A.~Awada, and R.~Youness, ``{A survey on quadrotors: Configurations,
  modeling and identification, control, collision avoidance, fault diagnosis
  and tolerant control},'' \emph{IEEE Aerospace and Electronic Systems
  Magazine}, vol.~33, no.~7, pp. 14--33, 2018.

\bibitem{guzman2019actuator}
J.~A. Guzm{\'{a}}n-Rabasa, F.~R. L{\'{o}}pez-Estrada, B.~M.
  Gonz{\'{a}}lez-Contreras, G.~Valencia-Palomo, M.~Chadli, and
  M.~Perez-Patricio, ``{Actuator fault detection and isolation on a quadrotor
  unmanned aerial vehicle modeled as a linear parameter-varying system},''
  \emph{Measurement and Control}, vol.~52, no. 9-10, pp. 1228--1239, 2019.

\bibitem{amoozgar2013experimental}
M.~H. Amoozgar, A.~Chamseddine, and Y.~Zhang, ``{Experimental test of a
  two-stage Kalman filter for actuator fault detection and diagnosis of an
  unmanned quadrotor helicopter},'' \emph{Journal of Intelligent {\&} Robotic
  Systems}, vol.~70, no.~1, pp. 107--117, 2013.

\bibitem{iannace2019fault}
G.~Iannace, G.~Ciaburro, and A.~Trematerra, ``{Fault diagnosis for UAV blades
  using artificial neural network},'' \emph{Robotics}, vol.~8, no.~3, p.~59,
  2019.

\bibitem{park2022multiclass}
J.~Park, Y.~Jung, and J.-H. Kim, ``{Multiclass Classification Fault Diagnosis
  of Multirotor UAVs Utilizing a Deep Neural Network},'' \emph{International
  Journal of Control, Automation and Systems}, vol.~20, no.~4, pp. 1316--1326,
  2022.

\bibitem{hinton2006reducing}
G.~E. Hinton and R.~R. Salakhutdinov, ``{Reducing the dimensionality of data
  with neural networks},'' \emph{science}, vol. 313, no. 5786, pp. 504--507,
  2006.

\bibitem{yang2021intelligent}
P.~Yang, H.~Geng, C.~Wen, and P.~Liu, ``{An Intelligent Quadrotor Fault
  Diagnosis Method Based on Novel Deep Residual Shrinkage Network},''
  \emph{Drones}, vol.~5, no.~4, p. 133, 2021.

\bibitem{yang2021intelligentb}
P.~Yang, C.~Wen, H.~Geng, and P.~Liu, ``{Intelligent fault diagnosis method for
  blade damage of quad-rotor UAV based on stacked pruning sparse denoising
  autoencoder and convolutional neural network},'' \emph{Machines}, vol.~9,
  no.~12, p. 360, 2021.

\bibitem{katta2022real}
S.~S. Katta, K.~Vuoj{\"{a}}rvi, S.~Nandyala, U.-M. Kovalainen, and L.~Baddeley,
  ``{Real-World On-Board Uav Audio Data Set For Propeller Anomalies},'' in
  \emph{ICASSP 2022-2022 IEEE International Conference on Acoustics, Speech and
  Signal Processing (ICASSP)}.\hskip 1em plus 0.5em minus 0.4em\relax IEEE,
  2022, pp. 146--150.

\bibitem{lecun1995convolutional}
Y.~LeCun, Y.~Bengio, and {others}, ``{Convolutional networks for images,
  speech, and time series},'' \emph{The handbook of brain theory and neural
  networks}, vol. 3361, no.~10, p. 1995, 1995.

\bibitem{vaswani2017attention}
A.~Vaswani, N.~Shazeer, N.~Parmar, J.~Uszkoreit, L.~Jones, A.~N. Gomez,
  L.~Kaiser, and I.~Polosukhin, ``{Attention is all you need},'' \emph{Advances
  in neural information processing systems}, vol.~30, 2017.

\bibitem{liu2020audio}
W.~Liu, Z.~Chen, and M.~Zheng, ``{An audio-based fault diagnosis method for
  quadrotors using convolutional neural network and transfer learning},'' in
  \emph{2020 American Control Conference (ACC)}.\hskip 1em plus 0.5em minus
  0.4em\relax IEEE, 2020, pp. 1367--1372.

\bibitem{krizhevsky2017imagenet}
A.~Krizhevsky, I.~Sutskever, and G.~E. Hinton, ``{Imagenet classification with
  deep convolutional neural networks},'' \emph{Communications of the ACM},
  vol.~60, no.~6, pp. 84--90, 2017.

\bibitem{Zhang2017}
W.~Zhang, G.~Peng, C.~Li, Y.~Chen, and Z.~Zhang, ``{A new deep learning model
  for fault diagnosis with good anti-noise and domain adaptation ability on raw
  vibration signals},'' \emph{Sensors (Switzerland)}, vol.~17, no.~2, 2017.

\bibitem{Zhang2018}
\BIBentryALTinterwordspacing
W.~Zhang, C.~Li, G.~Peng, Y.~Chen, and Z.~Zhang, ``{A deep convolutional neural
  network with new training methods for bearing fault diagnosis under noisy
  environment and different working load},'' \emph{Mechanical Systems and
  Signal Processing}, vol. 100, pp. 439--453, 2018. [Online]. Available:
  \url{http://dx.doi.org/10.1016/j.ymssp.2017.06.022}
\BIBentrySTDinterwordspacing

\bibitem{Srivastava2014}
N.~Srivastava, G.~Hinton, A.~Krizhevsky, I.~Sutskever, and R.~Salakhutdinov,
  ``{Dropout: A simple way to prevent neural networks from overfitting},''
  \emph{Journal of Machine Learning Research}, vol.~15, pp. 1929--1958, 2014.

\bibitem{Wen2018}
L.~Wen, X.~Li, L.~Gao, and Y.~Zhang, ``{A New Convolutional Neural
  Network-Based Data-Driven Fault Diagnosis Method},'' \emph{IEEE Transactions
  on Industrial Electronics}, vol.~65, no.~7, pp. 5990--5998, 2018.

\bibitem{tong2023machine}
J.~J. Tong, W.~Zhang, F.~Liao, C.~F. Li, and Y.~F. Zhang, ``{Machine Learning
  for UAV Propeller Fault Detection based on a Hybrid Data Generation Model},''
  \emph{arXiv preprint arXiv:2302.01556}, 2023.

\bibitem{Powers2015}
\BIBentryALTinterwordspacing
C.~Powers, D.~Mellinger, and V.~Kumar, ``{Quadrotor Kinematics and Dynamics},''
  in \emph{Handbook of Unmanned Aerial Vehicles}, K.~P. Valavanis and G.~J.
  Vachtsevanos, Eds.\hskip 1em plus 0.5em minus 0.4em\relax Dordrecht: Springer
  Netherlands, 2015, pp. 307--328. [Online]. Available:
  \url{https://doi.org/10.1007/978-90-481-9707-1_71}
\BIBentrySTDinterwordspacing

\bibitem{van2008visualizing}
L.~der Maaten and G.~Hinton, ``{Visualizing data using t-SNE.}'' \emph{Journal
  of machine learning research}, vol.~9, no.~11, 2008.

\end{thebibliography}

%



\end{document}